\documentclass[conference]{IEEEtran}
\IEEEoverridecommandlockouts

\usepackage[utf8]{inputenc}  
\usepackage[T1]{fontenc}     
\usepackage{url}           
\usepackage{booktabs}        
\usepackage{amsfonts}        
\usepackage{nicefrac}        
\usepackage{microtype}       
\usepackage{yhmath}         
\usepackage{todonotes}

\usepackage{multirow}
\usepackage{hhline}
\usepackage{color}
\usepackage{epsfig}
\usepackage{subfig}
\usepackage{graphicx}
\usepackage{amsmath}
\usepackage{amssymb}
\usepackage{amsthm}
\usepackage{bm}
\usepackage{algorithm}
\usepackage{algorithmicx}
\usepackage[noend]{algpseudocode}
\usepackage{comment}
\usepackage[numbers,sort&compress]{natbib}
\setcitestyle{numbers,square,comma}
\usepackage{epstopdf}
\usepackage{physics}
\usepackage{setspace}
\usepackage{cleveref}
\usepackage{subfig}

\graphicspath{{./figures/}}

\definecolor{officegreen}{rgb}{0.0, 0.5, 0.0}

\newtheorem{theorem}{Theorem}

\newtheorem{assumption}[theorem]{Assumption}

\def\R{\mathbb{R}}

\def\BS{\bm{S}}
\def\BC{{\bm{C}}}
\def\BU{{\bm{U}}}
\def\BR{{\bm{R}}}
\def\BV{\bm{V}}
\def\BW{\bm{W}}
\def\BSigma{\bm{\Sigma}}
\def\BX{\bm{X}}
\def\BY{\bm{Y}}

\def\BM{\bm{M}}
\def\BW{\bm{W}}
\def\cI{{\mathcal{I}}}
\def\cJ{{\mathcal{J}}}
\def\cT{{\mathcal{T}}}
\def\cD{\mathcal{D}}

\def\cO{\mathcal{O}}
\def\cP{\mathcal{P}}
\def\be{\bm{e}}

\def\fro{{\mathrm{F}}}
\def\rank{\mathrm{rank}}

\DeclareMathOperator*{\minimize}{\mathrm{minimize}}
\DeclareMathOperator*{\subject}{\mathrm{subject~to}}
\def\poly{\mathrm{poly}}
\def\polylog{\mathrm{polylog}}

\def\BibTeX{{\rm B\kern-.05em{\sc i\kern-.025em b}\kern-.08em
    T\kern-.1667em\lower.7ex\hbox{E}\kern-.125emX}}
\begin{document}

\title{On the Robustness of Cross-Concentrated Sampling  for Matrix Completion\\
\thanks{The authors contributed equally. This work was partially supported by NSF DMS 2304489.
(Corresponding authors: HanQin Cai and Longxiu Huang.)}
}

\author{
\IEEEauthorblockN{HanQin Cai}
\IEEEauthorblockA{\textit{Department of Statistics and Data Science} \\ \textit{Department of Computer Science} \\
\textit{University of Central Florida}\\
 Orlando, FL 32816, USA\\
hqcai@ucf.edu}
\and
\IEEEauthorblockN{Longxiu Huang}
\IEEEauthorblockA{\textit{Department of Computational Mathematics, Science and Engineering } \\
\textit{Department of Mathematics} \\
\textit{ Michigan State University}\\
 East Lansing, MI 48824, USA \\
huangl3@msu.edu}
\and
\IEEEauthorblockN{Chandra Kundu}
\IEEEauthorblockA{\textit{Department of Statistics and Data Science} \\ 
\textit{University of Central Florida}\\
 Orlando, FL 32816, USA\\
chandra.kundu@ucf.edu}
\and
\IEEEauthorblockN{Bowen Su}
\IEEEauthorblockA{
\textit{Department of Mathematics}\\
\textit{ Michigan State University}\\
 East Lansing, MI 48824, USA \\
subowen@msu.edu}
}

\maketitle

\begin{abstract}
Matrix completion is one of the crucial tools in modern data science research. 
Recently, a novel sampling model for matrix completion coined cross-concentrated sampling (CCS) has caught much attention. However, the robustness of the CCS model against sparse outliers remains unclear in the existing studies. In this paper, we aim to answer this question by exploring a novel Robust CCS Completion problem. A highly efficient non-convex iterative algorithm, dubbed Robust CUR Completion (RCURC), is proposed. The empirical performance of the proposed algorithm, in terms of both efficiency and robustness, is verified in synthetic and real datasets.
\end{abstract}


\section{Introduction}
\IEEEPARstart{M}{atrix} completion problem \cite{candes2009exact,recht2011simpler} aims to recover the underlying low-rank matrix $\BX$ from some entry-wise partial observation. It has received tremendous attention in the past decades and has been widely applied in real-world applications such as collaborative filtering \cite{Netflix,goldberg1992using}, image processing \cite{chen2004recovering,hu2012fast}, and signal processing \cite{cai2019fast,cai2023hsgd}. 
While the vanilla matrix completion studies are often based on entry-wise uniform or Bernoulli sampling models, the recent development of a novel sampling model called \textit{Cross-Concentrated Sampling} (CCS) \cite{cai2023ccs} has caught the attention of the matrix completion community. The CCS model is summarized in Procedure~\ref{alg:CCS}. 

\floatname{algorithm}{Procedure}
\begin{algorithm}[ht]
\caption{Cross-Concentrated Sampling (CCS) \cite{cai2023ccs}}\label{alg:CCS}
\begin{algorithmic}[1]
\State \textbf{Input:}  $\BY$: access to the data matrix. 
\State Uniformly choose row and column indices $\cI,\cJ$.
\State Set $\BR=[\BY]_{\cI,:}$ and $\BC=[\BY]_{:,\cJ}$.
\State Uniformly sample entries in $\BR$ and $\BC$, then record the sampled locations as $\Omega_{\BR}$ and $\Omega_{\BC}$, respectively.
\State\textbf{Output:}  $[\BY]_{\Omega_{\BR}\cup\Omega_{\BC}}$, $\Omega_{\BR}$, $\Omega_{\BC}$, $\cI$, $\cJ$.
\end{algorithmic}
\end{algorithm}
\floatname{algorithm}{Algorithm}

\begin{figure*}[h]
    \centering
     \subfloat[Uniform Sampling]{\includegraphics[width=0.24\textwidth]{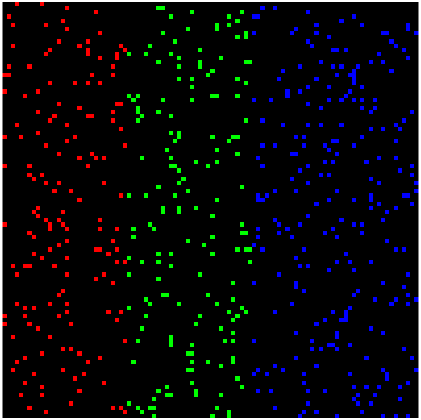}
    }
    \hfill
    \subfloat[CCS--Less Concentrated]{\includegraphics[width=0.24\textwidth]{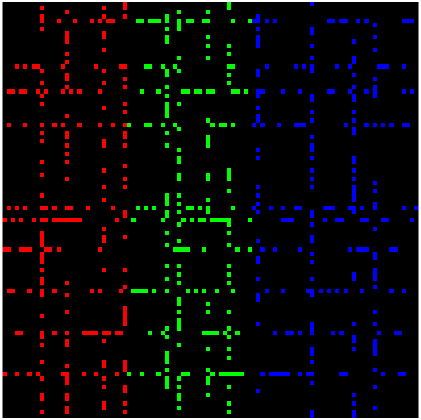}
    }
    \hfill
    \subfloat[CCS--More Concentrated]{\includegraphics[width=0.24\textwidth]{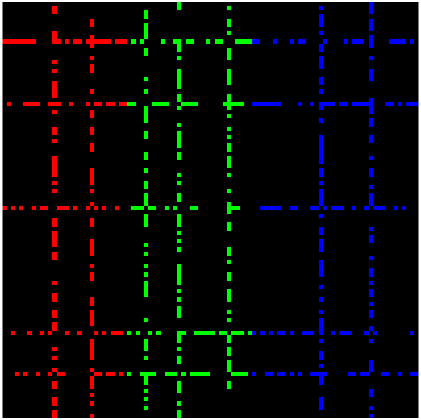}
    }\hfill
      \subfloat[CUR Sampling]{\includegraphics[width=0.24\textwidth]{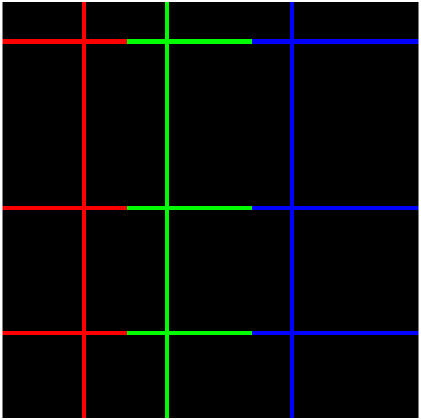}
    }\\
    \caption{\cite{cai2023ccs} Visual illustrations of different sampling schemes. From left to right, sampling methods change from the uniform sampling style to the CUR sampling style with the same total observation rate. Colored pixels indicate observed entries, while black pixels mean missing entries.} \label{fig:CCS}
\end{figure*}

As illustrated in \Cref{fig:CCS}, the CCS model bridges two popular sampling models for matrix completion: \textit{Uniform sampling} and \textit{CUR sampling} which recovers the underlying low-rank matrix from observed rows and columns. The CCS model provides extra flexibility, with a theoretical guideline, for samples concentrated on certain rows and columns. This model has the potential for more cost- and time-efficient sampling strategies in applications such as recommendation systems. However, data in real-world applications are often corrupted by sparse outliers, and uniformly sampled data must be processed by some robust completion algorithms for reliable recovery results. That being said, a crucial question has to be answered before the CCS model can be widely used: \textit{``Is CCS-based matrix completion robust to sparse outliers under some robust algorithms, like the uniform sampling model?''} 

Mathematically, provided the partial observation $\cP_\Omega(\BY)$ of a corrupted data matrix $\BY=\BX+\BS$, we study the Robust CCS Completion problem:
\begin{equation} \label{eq:robust ccs}
    \begin{split}
     \minimize_{\BX,\BS}&\quad\frac{1}{2}\langle\cP_{\Omega}(\BX+\BS-\BY),\BX+\BS-\BY\rangle\\ 
     \subject&\quad \rank(\BX)=r \quad \textnormal{and}\quad \BS \textnormal{ is $\alpha$-sparse,}
    \end{split}
\end{equation} 
where $\cP$ is the sampling operator defined as: 
\begin{equation*} \label{eq:sample operator}
\cP_{\Omega}(\BY)=\sum_{(i,j)\in\Omega}[\BY]_{i,j}\be_i\be_j^\top
\end{equation*}
and the observation set $\Omega$ is generated according to the CCS model. Similar to the standard robust matrix completion, 
a sparse variable $\BS$ is introduced for better outlier tolerance, thus one can recover the underlying low-rank matrix $\BX$ accurately.

\subsection{Related Work}
The problem of low-rank matrix recovery in the presence of sparse outliers has been well-studied under the settings of uniform sampling and Bernoulli sampling. This problem is known as \textit{robust principal component analysis} (RPCA) when the corrupted data matrix is fully observed, and it is called \textit{robust matrix completion} (RMC) when data is partially observed. The seminal work \cite{candes2011robust} considers both RPCA and RMC problems via convex relaxed formulations and provides recovery guarantees. In particular, under the $\mu$-incoherence assumption for the low-rank $\BX$, \cite{candes2011robust} requires the positions of outliers to be placed uniformly at random, and at least $0.1n^2$ entries are observed uniformly at random. 
Later, a series of non-convex algorithms \cite{
netrapalli2014non,yi2016fast,zhang2018robust,cai2019accelerated,tong2021accelerating,cai2021lrpca} tackle RPCA and/or RMC problems with an improved, non-random $\alpha$-sparsity assumptions for the outlier matrix $\BS$. The typical recovery guarantee shows a linear  convergence of a non-convex algorithm, provided $\alpha\leq\cO(1/\poly(\mu r))$; moreover, $\cO(\poly(\mu r)\polylog(n)n)$ random samples are typically required for the RMC cases. 
Another line of work \cite{chen2014robust,cai2019fast,zhang2019correction,cai2021accelerated,cai2023hsgd} focuses on the robust recovery of structured low-rank matrices, e.g., Hankel matrices, and they typically require merely $\cO(\poly(\mu r)\polylog(n))$ samples by utilizing the structure, even in the presence of structured outliers.
More recently, \cite{cai2021ircur,cai2021rcur,hamm2022RieCUR} study the robust CUR decomposition problem, that is, recovering the low-rank matrix from row- and column-wise observations with entry-wise corruptions. The studies show that robust CUR decomposition is as robust as RPCA, with better computational efficiency and data interoperability. 

On the other hand, \cite{cai2023ccs} shows that CCS-based matrix completion requires $\cO(\mu^2r^2 n\log^2 n)$ samples which is only a factor of $\log n$ worse than the state-of-the-art result; however, its outlier tolerance has not been studied.

\subsection{Notation}
For a matrix $\BM$, $[\BM]_{i,j}$, $[\BM]_{\cI,:}$, $[\BM]_{:,\cJ}$, and $[\BM]_{\cI,\cJ}$ denote its $(i,j)$-th entry,  its row submatrix with row indices $\cI$, its column submatrix with column indices $\cJ$, and its submatrix with row indices $\cI$ and column indices $\cJ$, respectively.  $\|\BM\|_{\fro}:=(\sum_{i,j}[\BM]_{i,j}^2)^{1/2}$ denotes the Frobenius norm, $\|\BM\|_{2,\infty}:=\max_{i}(\sum_{j}[\BM]_{i,j}^2)^{1/2}$ denotes the largest row-wise $\ell_2$-norm, $\|\BM\|_{\infty}=\max_{i,j}|[\BM]_{i,j}|$ denote the largest entry-wise magnitude,  $\BM^\dagger$ represents the Moore–Penrose inverse of $\BM$, and $\BM^\top$ is the transpose of $\BM$. 
$\langle\cdot,\cdot\rangle$ denotes the Frobenius inner product. 
We denote $\bm{e}_i$ as the $i$-th standard basis of the real vector space. 
The symbol $[n]$ denotes the set $\{1,2,\cdots,n\}$ for all $n\in\mathbb{Z}^+$. 
Throughout the paper, uniform sampling is referred to as uniform sampling with replacement.

\section{Preliminaries} \label{sec:preliminaries}

\subsection{Assumptions}
As discussed in the related work, $\mu$-incoherence and $\alpha$-sparsity are commonly used assumptions in RMC problems. We will use the same assumptions in this study and give the formal definition here:

\begin{assumption}[$\mu$-incoherence] \label{as:incoherence}
Let $\BX\in\R^{n_1 \times n_2}$ be a rank-$r$ matrix. It is said $\mu$-incoherent if
\begin{gather*}
    \|\BU\|_{2,\infty}\leq \sqrt{\frac{\mu r}{n_1}} 
    \qquad\textnormal{and}\qquad
    \|\BV\|_{2,\infty}\leq \sqrt{\frac{\mu r}{n_2}}
\end{gather*}
for some positive $\mu$, where $\BU\BSigma\BV^\top$ is the compact SVD of $\BX$. 
\end{assumption}

\begin{assumption}[$\alpha$-sparsity] \label{as:sparsity}
$\BS\in\R^{n_1 \times n_2}$ is said $\alpha$-sparse if there are at most $\alpha$ fraction of non-zero entries in each of its row and column. That is, for all $i\in[n_1]$ and $j\in[n_2]$, it holds
\begin{gather*}
    \|[\BS]_{i,:}\|_0\leq\alpha n_2
    \qquad\textnormal{and}\qquad
    \|[\BS]_{:,j}\|_0\leq\alpha n_1.
\end{gather*}
Note that no randomness is required for the sparsity pattern.
\end{assumption}

\subsection{CUR Approximation}

CUR approximation, also known as skeleton decomposition, is the backbone of our algorithm design. We shall go over some background of CUR approximation for the reader's interest. 
CUR approximation is a self-expressive technique that fills the interpretability gap in matrix dimensionality reduction. Given  a rank-$r$ matrix $\BX\in \R^{n \times n}$, if we select rows and columns from $\BX$ that span its row and column spaces respectively, then $\BX$ can be reconstructed from these submatrices. This has been affirmed in previous studies:

\begin{theorem}[\citep{HH2020}]\label{thm:noiseless CUR}
Consider row and column index sets $\cI,\cJ\subseteq[n]$.  Denote submatrices $\BC=[\BX]_{:,\cJ}$, $\BU=[\BX]_{\cI,\cJ}$ and $\BR=[\BX]_{\cI,:}$.  If $\rank(\BU)=\rank(\BX)$, then 
$\BX = \BC \BU^\dagger \BR$.
\end{theorem}

For further details of CUR approximation, including the historical context and formal proof, one can refer to \citep{HH2020} and reference therein. Note that through various row- and column-wise sampling methodologies, meeting this rank equivalence condition in \Cref{thm:noiseless CUR} is highly probable, especially when a sufficient number of rows and columns are sampled. An example of this is detailed in \Cref{thm:uniform sample}. 

\begin{theorem}[{\citep[Theorem~1.1]{chiu2013sublinear}}] \label{thm:uniform sample}
    Let $\BX$ satisfy \Cref{as:incoherence}, and suppose we sample $|\cI|=\cO(r \log n)$ rows and $|\cJ|=\cO(r \log n)$ columns uniformly at random. Then $\rank(\BU)=\rank(\BX)$ with probability at least $1-\mathcal{O}(r n^{-2})$.
\end{theorem}

There also exist many other sampling methodologies that ensure the recoverability of CUR approximation;  for instance,  the Discrete Empirical Interpolation Method (DEIM) \cite{chaturantabut2010nonlinear,SorensenDEIMCUR}  and leverage scores sampling method \cite{DMM08}. The interested reader can find more details in \cite[Table 1]{HH2020S}. Nonetheless, they all sample completed rows and columns, thus are substantially different than the CCS model.




\section{Proposed Algorithm} \label{sec:algorithm}
In this section, we will present a novel non-convex algorithm for solving the proposed Robust CCS Completion problem \eqref{eq:robust ccs}. This iterative algorithm is based on the idea of projected gradient descent where CUR approximation is used for fast low-rank approximation in each iteration. The algorithm is dubbed Robust CUR Completion (RCURC) and summarized in \Cref{alg:RCURC}. 

\begin{algorithm}[h]
\caption{Robust CUR Completion (RCURC)}  \label{alg:RCURC}
\begin{algorithmic}[1]
\State \textbf{Input:} 
$[\BY=\BX+\BS]_{\Omega_{\BR}\cup\Omega_{\BC}}$: observed data; 
$\Omega_\BR, \Omega_\BC$: observation locations; 
$\cI,\cJ$: row and column indices that define $\BR$ and $\BC$ respectively; $\eta_R,\eta_C$: step sizes; 
$r$: target rank;
$\varepsilon$: target precision level;
$\zeta_0$: initial thresholding value;
$\gamma$: thresholding decay parameter.

\State $\BX_0=\bm{0}$; \quad $k=0$

\While {$e_k > \varepsilon$} \algorithmiccomment{\textcolor{officegreen}{$e_k$ is defined in \eqref{eq:e_k}}}
    \State \textcolor{officegreen} {// Updating sparse component}
    \State $\zeta_{k+1} = \gamma^k\zeta_0$
    \State $[\BS_{k+1}]_{\cI,:} = \cT_{\zeta_{k+1}}[\BY-\BX_k]_{\cI,:}$
    \State $[\BS_{k+1}]_{:,\cJ} = \cT_{\zeta_{k+1}}[\BY-\BX_k]_{:,\cJ}$
    
    \State \textcolor{officegreen} {// Updating low-rank component}
    \State $\BR_{k+1}=[\BX_k]_{\cI,:}+\eta_R[\cP_{\Omega_\BR}(\BY-\BX_k-\BS_{k+1})]_{\cI,:}$
    \State $\BC_{k+1}=[\BX_k]_{:,\cJ}+\eta_C[\cP_{\Omega_\BC}(\BY-\BX_k-\BS_{k+1})]_{:,\cJ}$
    \State $\BU_{k+1}=\cD_r([\BR_{k+1}]_{:,\cJ}\uplus[\BC_{k+1}]_{\cI,:})$
    \State $[\BR_{k+1}]_{:,\cJ}=\BU_{k+1}$ \quad \textnormal{and} \quad $[\BC_{k+1}]_{\cI,:}=\BU_{k+1}$
    \State $\BX_{k+1} = \BC_{k+1}\BU_{k+1}^{\dagger}\BR_{k+1}$ \algorithmiccomment{\textcolor{officegreen}{Do not compute}}
    \State $k = k + 1$
\EndWhile
\State \textbf{Output:} 
$\BC_k, \BU_k, \BR_k$: The CUR components of the estimated low-rank matrix.
\end{algorithmic}
\end{algorithm} 

We will go over the algorithm step by step in the following paragraphs. For the ease of the presentation, we start with the low-rank component. 

\vspace{0.1in}
\noindent\textbf{Updating low-rank component.} To utilize the structure of cross-concentrated samples, it is efficient to enforce the low-rank constraint of $\BX$ with the CUR approximation technique. Let $\BR=[\BX]_{\cI,:}$, $\BC=[\BX]_{:,\cJ}$, and $\BU=[\BX]_{\cI,\cJ}$. Applying gradient descent directly on $\BR$ and $\BC$ gives: 
\begin{equation*}
    \begin{split}
    \BR_{k+1}&=[\BX_k]_{\cI,:}+\eta_R[\cP_{\Omega_\BR}(\BY-\BX_k-\BS_{k+1})]_{\cI,:} \\
    \BC_{k+1}&=[\BX_k]_{:,\cJ}+\eta_C[\cP_{\Omega_\BC}(\BY-\BX_k-\BS_{k+1})]_{:,\cJ},
    \end{split}
\end{equation*}
where $\eta_R$ and $\eta_C$ are the stepsizes. However, When it comes to the intersection submatrix $\BU$, it is more complicated as $\Omega_R$ and $\Omega_C$ can have overlaps. We abuse the notation $\uplus$ and define an operator called \textit{union sum} here: 
\begin{equation*}
\begin{split}
    &~\big[[\BR_{k+1}]_{:,\cJ}\uplus[\BC_{k+1}]_{\cI,:}\big]_{i,j}\\
    =&\begin{cases}
        [\BR_{k+1}]_{i,j} & \textnormal{if } (i,j)\in\Omega_R\setminus\Omega_C;\\
        [\BC_{k+1}]_{i,j} & \textnormal{if } (i,j)\in\Omega_C\setminus\Omega_R;\\
        \frac{\eta_R\eta_C}{\eta_R+\eta_C}\left(\frac{[\BR_{k+1}]_{i,j}}{\eta_R}+\frac{[\BC_{k+1}]_{i,j}}{\eta_C}\right) & \textnormal{if } (i,j)\in\Omega_C\cap\Omega_R;\\
        0 & \textnormal{otherwise}.
    \end{cases}
\end{split}
\end{equation*}
Basically, we take whatever value we have for the non-overlapped entries and take a weighted average for the overlaps where the weights are determined by the stepsizes used in the updates of $\BR_{k+1}$ and $\BC_{k+1}$. To ensure the rank-$r$ constraint, at least one of $\BC_{k+1}$, $\BU_{k+1}$ or $\BR_{k+1}$ should be rank-$r$. For computational efficiency, we choose to put it on the smallest one. Thus, 
\begin{equation*}
    \BU_{k+1}=\cD_r\left([\BR_{k+1}]_{:,\cJ}\uplus[\BC_{k+1}]_{\cI,:}\right),
\end{equation*}
where $\cD_r$ is the best rank-$r$ approximation operator via truncated SVD. After replacing the intersection part $\BU_{k+1}$ in the previously updated $\BR_{k+1}$ and $\BC_{k+1}$, we have the new estimation of low-rank component: 
\begin{equation} \label{eq:update X}
    \BX_{k+1} = \BC_{k+1}\BU_{k+1}^{\dagger}\BR_{k+1}.
\end{equation}
However, \eqref{eq:update X} is just a conceptual step and one should never compute it. In fact, the full matrix $\BX$ is never needed and should not be formed in the algorithm as updating the corresponding CUR components is sufficient.  

\vspace{0.1in}
\noindent\textbf{Updating sparse component.} 
We detect the outliers and put them into the sparse matrix $\BS$ via hard-thresholding operator:
\begin{equation*}
    [\cT_\zeta(\BM)]_{i,j}=
    \begin{cases}
        0 & \textnormal{if } |[\BM]_{i,j}|<\zeta; \\
        [\BM]_{i,j} & \textnormal{otherwise.}
    \end{cases}
\end{equation*}
The hard-thresholding on residue $\BY-\BX_k$, paired with iterative decayed thresholding values:
\begin{equation*}
    \zeta_{k+1} = \gamma^k\zeta_0 \quad\textnormal{with some } \gamma\in(0,1),
\end{equation*}
has shown promising performance in outlier detection in prior art \cite{cai2019accelerated,cai2021lrpca,cai2021rtcur,cai2024rtcur}. Notice that we only need to remove outliers located on the selected rows and columns, i.e., $\BR$ and $\BC$, since they are the only components needed to update the low-rank component later. Therefore, for computational efficiency, we should only compute $\BX_k$ on the selected rows and columns to update $\BS_{k+1}$ correspondingly---as said, one should never compute the full $\BX_k$ in this algorithm. In particular,
\begin{align*}
    [\BX_k]_{\cI,:} =  [\BC_{k}]_{\cI,:}\BU_{k}^\dagger\BR_{k}    \textnormal{~~and~~} 
    [\BX_k]_{:,\cJ} =  \BC_{k}\BU_{k}^\dagger[\BR_{k}]_{:,\cJ}.
\end{align*}

\vspace{0.1in}
\noindent\textbf{Stopping criteria and stepsizes.} We finish the algorithm with the last few pieces. The stopping criteria is set to be $e_k\leq \varepsilon$ where $\varepsilon$ is the targeted accuracy and the computational error is 
\begin{equation} \label{eq:e_k}
    e_k=\frac{\langle\cP_{\Omega_\BR\cup\Omega_\BC}(\BS_k+\BX_k-\BY),\BS_k+\BX_k-\BY\rangle}{\langle\cP_{\Omega_\BR\cup\Omega_\BC}\BY,\BY\rangle}.
\end{equation}
The recommended stepsizes are $\eta_R=\frac{1}{p_R}$ and $\eta_C = \frac{1}{p_C}$ where $p_R$ and $p_C$ are the observation rates of $\Omega_R$ and $\Omega_C$ respectively. Smaller stepsizes should be used with larger $\alpha$, i.e., more outliers.

\section{Numerical Experiments} \label{sec:numerical}
In this section, we will verify the empirical performance of RCURC with both synthetic and real datasets. 
All experiments are implemented on Matlab R2022b and executed on a laptop equipped with  Intel i7-11800H CPU (2.3GHz @ 8 cores) and 16GB DDR4 RAM.

\subsection{Synthetic Datasets}
 
\begin{figure}[ht!]
\includegraphics[width=.49\linewidth]{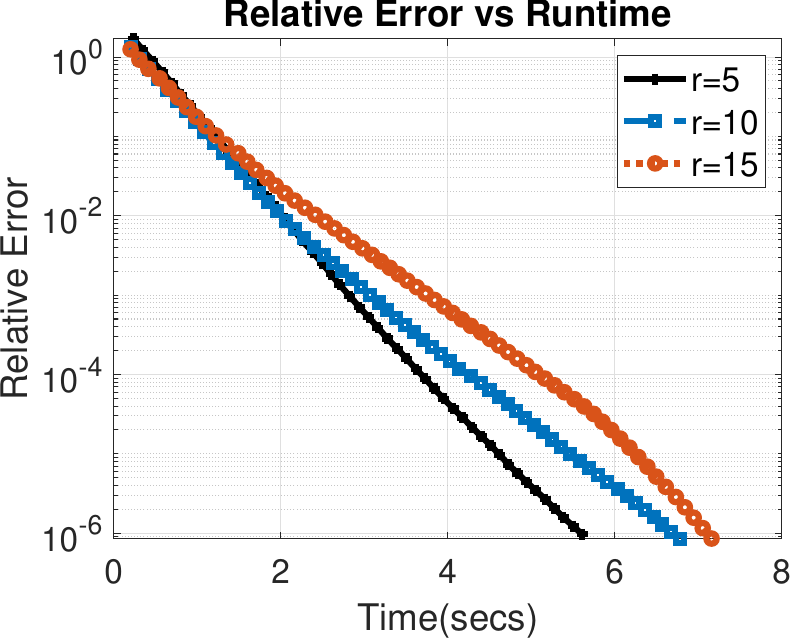}
\includegraphics[width=0.49\linewidth]{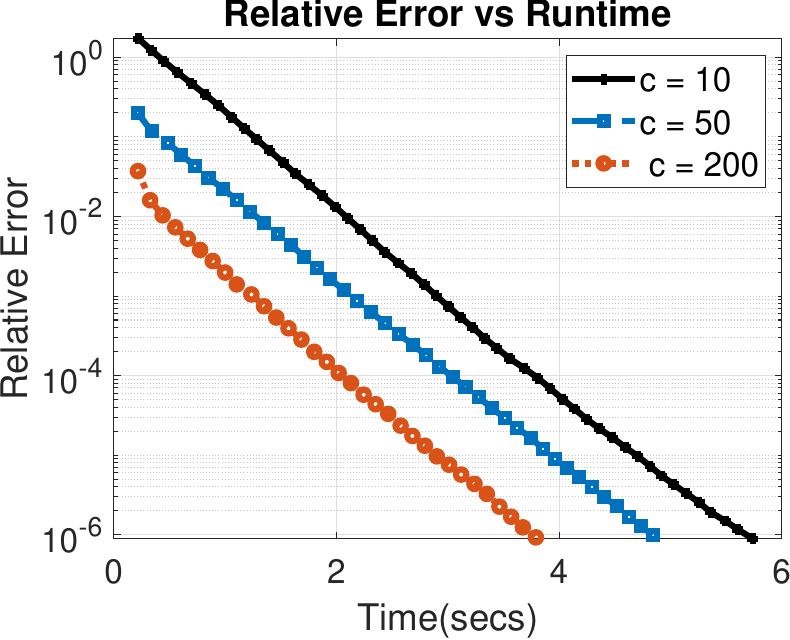}
\caption{\small Empirical convergence of RCURC. \textbf{Left:} $c=10$ and varying $r$. 
 \textbf{Right:} $r=5$ and varying $c$. }\label{conv_RCURC}
 \end{figure}

In this simulation, we assess the computational efficiency of our algorithm, RCURC, in addressing the robust CCS completion problem. We construct \(\BY = \BX + \BS\), a \(d \times d\) matrix with $d=3000$, where \(\BX = \BW\BV^\top\) is a randomly generated rank-\(r\) matrix. To create the sparse outlier tensor \(\BS\), we randomly select \(\alpha\) percent entries to form the support of \(\BS\). The values of the non-zero entries are then uniformly sampled from the range \([-c\mathbb{E}(\left|\BS_{ij}\right|), c\mathbb{E}(\left|\BS_{ij}\right|)]\). To generate the robust CCS completion problems, we set  $\alpha=0.2$, $\frac{|\cI|}{d}=\frac{|\cJ|}{d}=30\%$, and $\frac{|\Omega_\BR|}{|\cI|d}=\frac{|\Omega_\BC|}{|\cJ|d}=25\%$. The results are obtained by averaging over $50$ runs and reported in \Cref{conv_RCURC}. 
Both figures in \Cref{conv_RCURC} depict the relationship between the relative error \(e_k\) and computational time for our RCURC method with varying rank \(r\) and outlier amplification factor \(c\). It is noteworthy that RCURC consistently achieves nearly linear convergence rates across different scenarios.

\subsection{Video Background Subtraction}
We have applied our robust CCS model and RCURC to the problem of background separation. Two popular videos, \textit{shoppingmall} and  \textit{restaurant}, are used as benchmarks.
The dimensions of the datasets are  documented in Table~\ref{tab:video}.  

\begin{figure}[ht!]
\vspace{-0.02in}
\centering
\subfloat{\includegraphics[width=.245\linewidth]{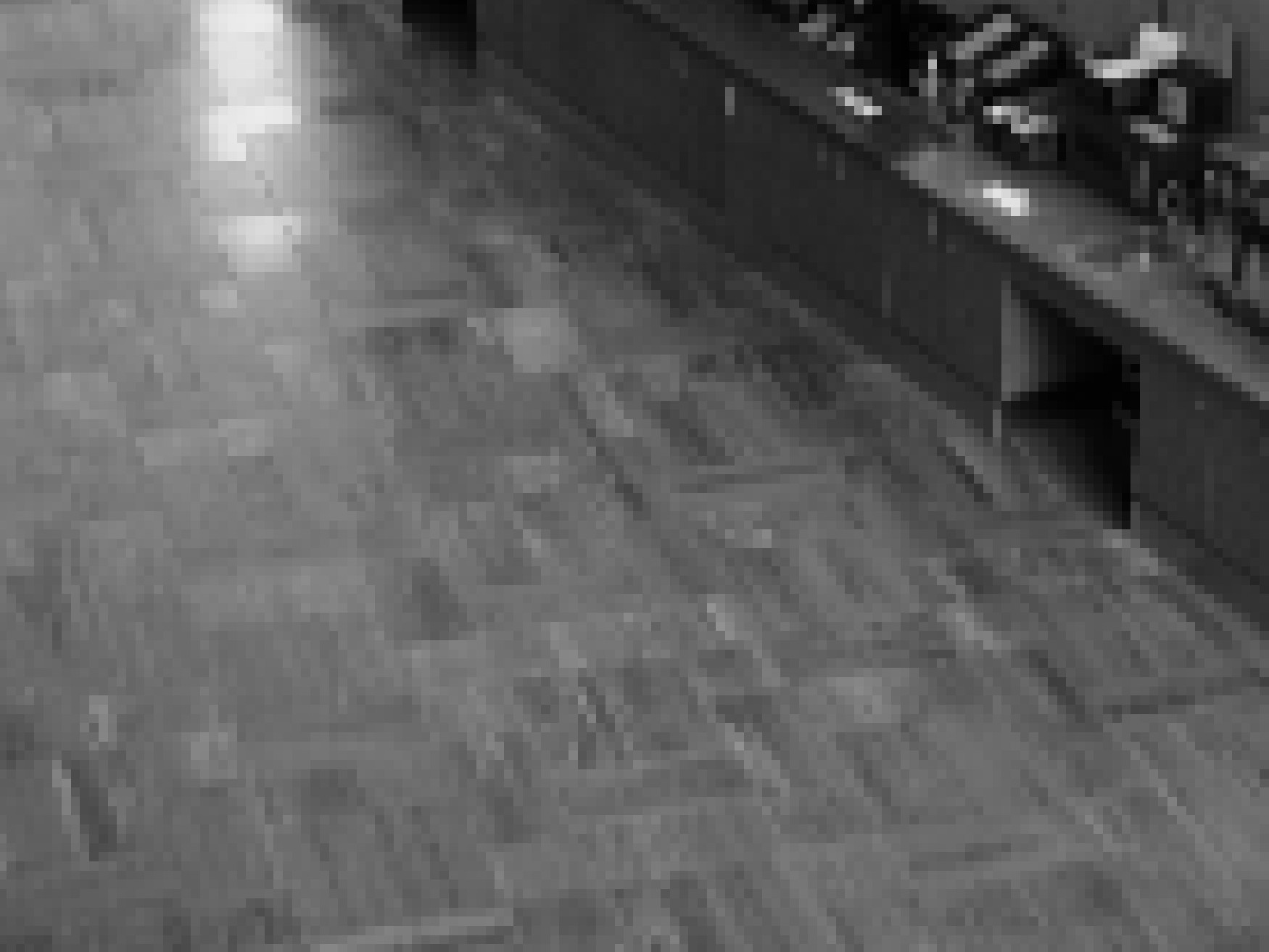}}\hfill
\subfloat{\includegraphics[width=.245\linewidth]{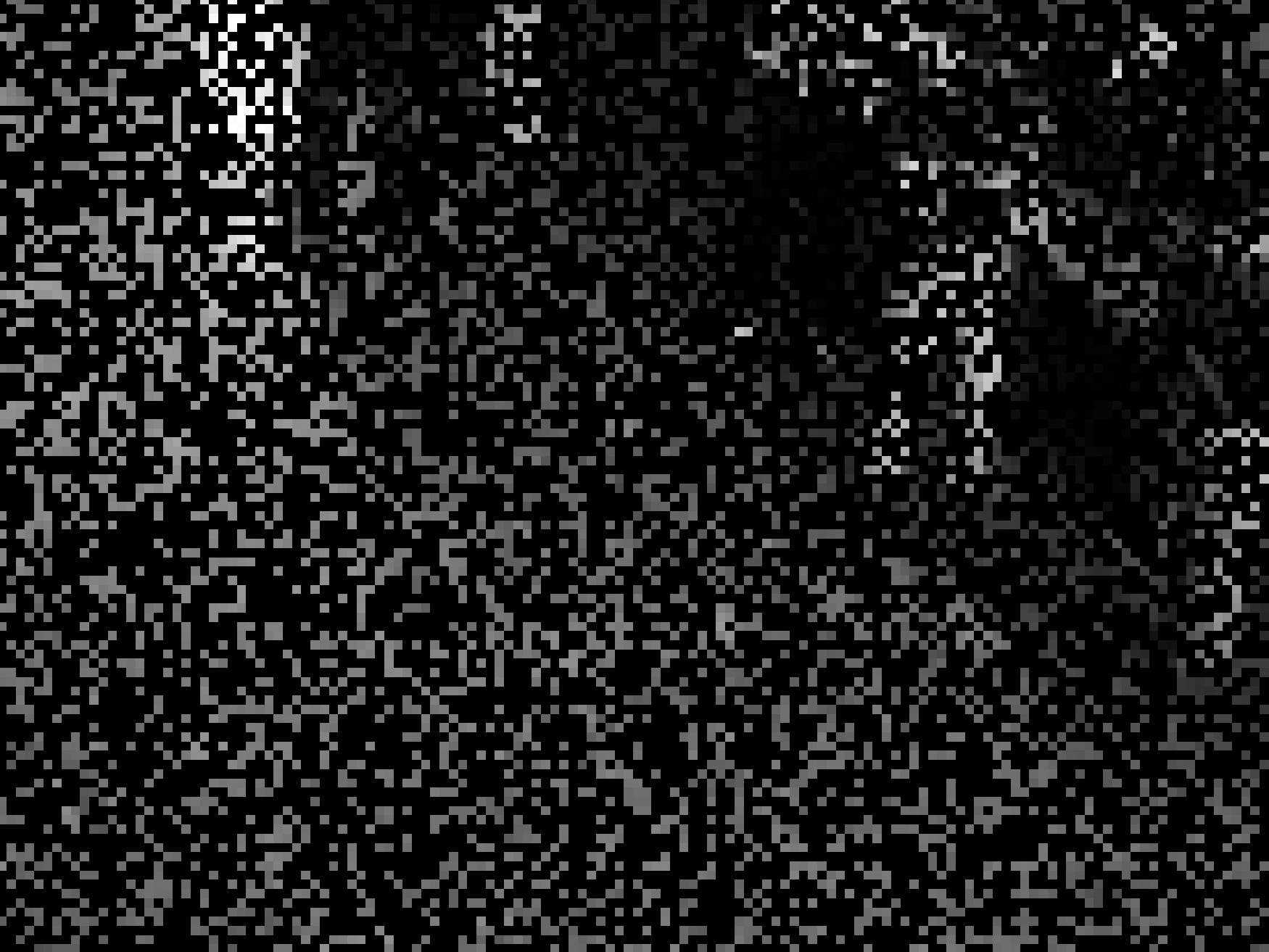}} \hfill
\subfloat{\includegraphics[width=.245\linewidth]{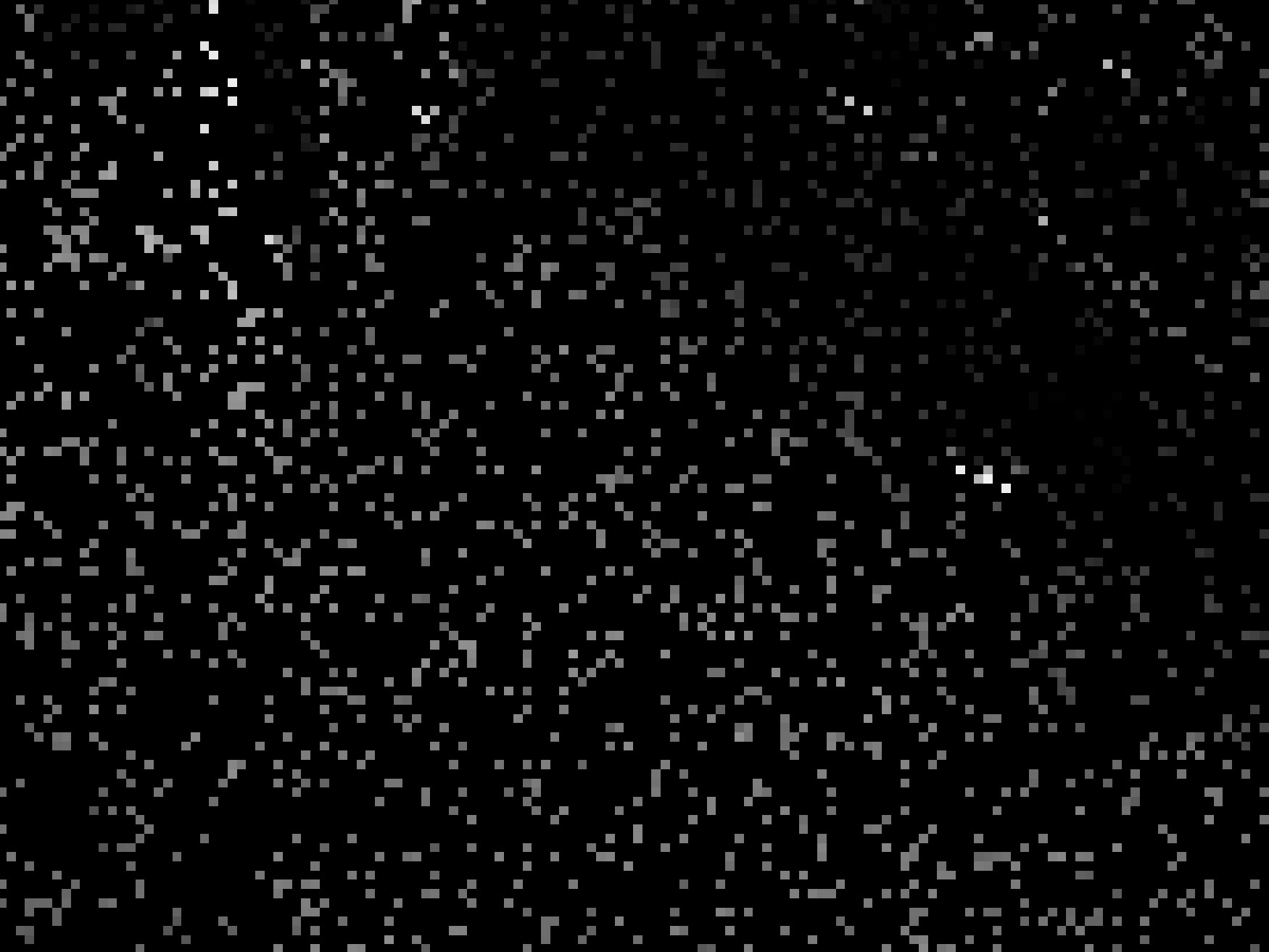}}\hfill
\subfloat{\includegraphics[width=.245\linewidth]{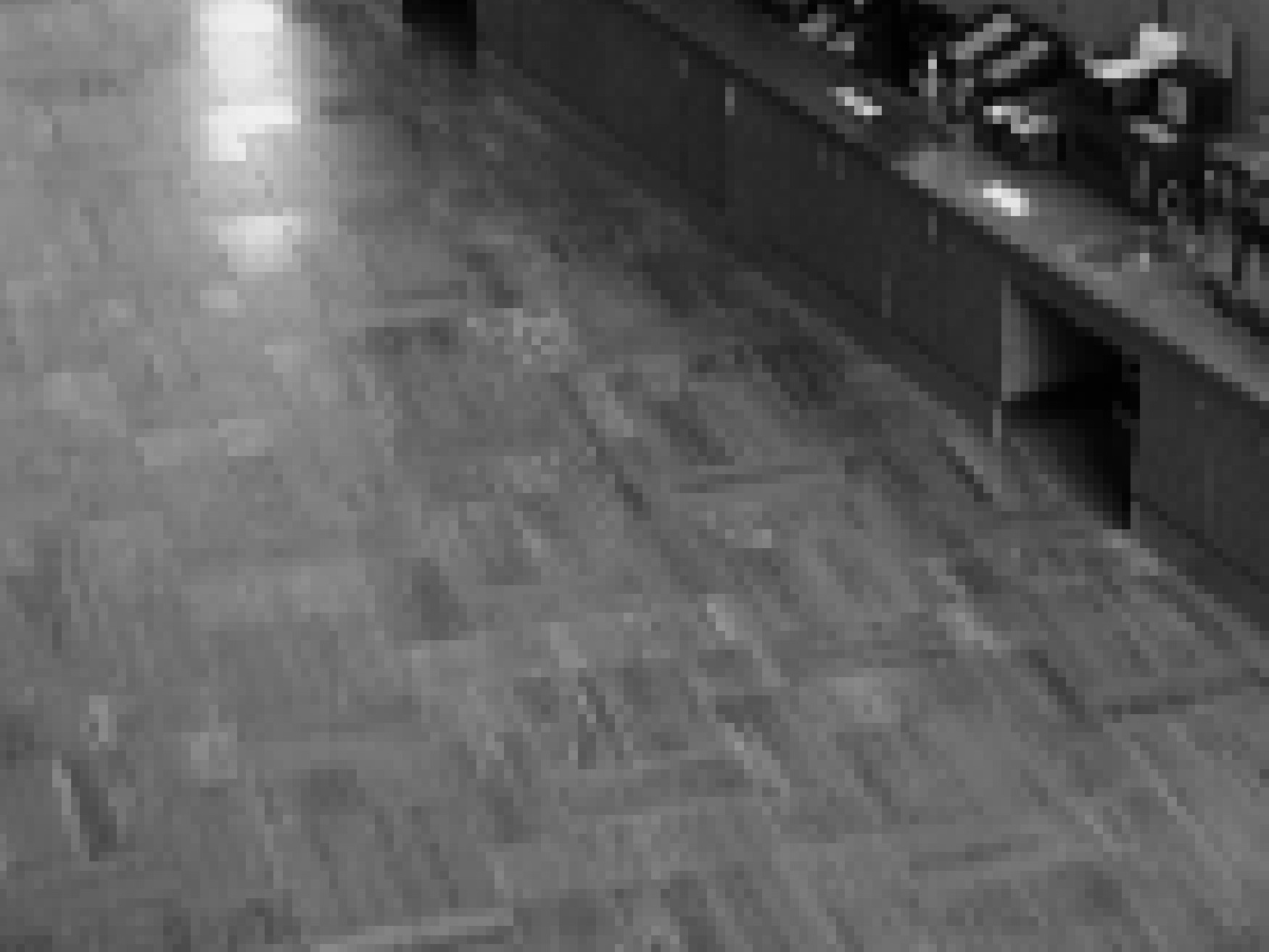}} 
 \vspace{-0.12in}
 
\subfloat{\includegraphics[width=.245\linewidth]{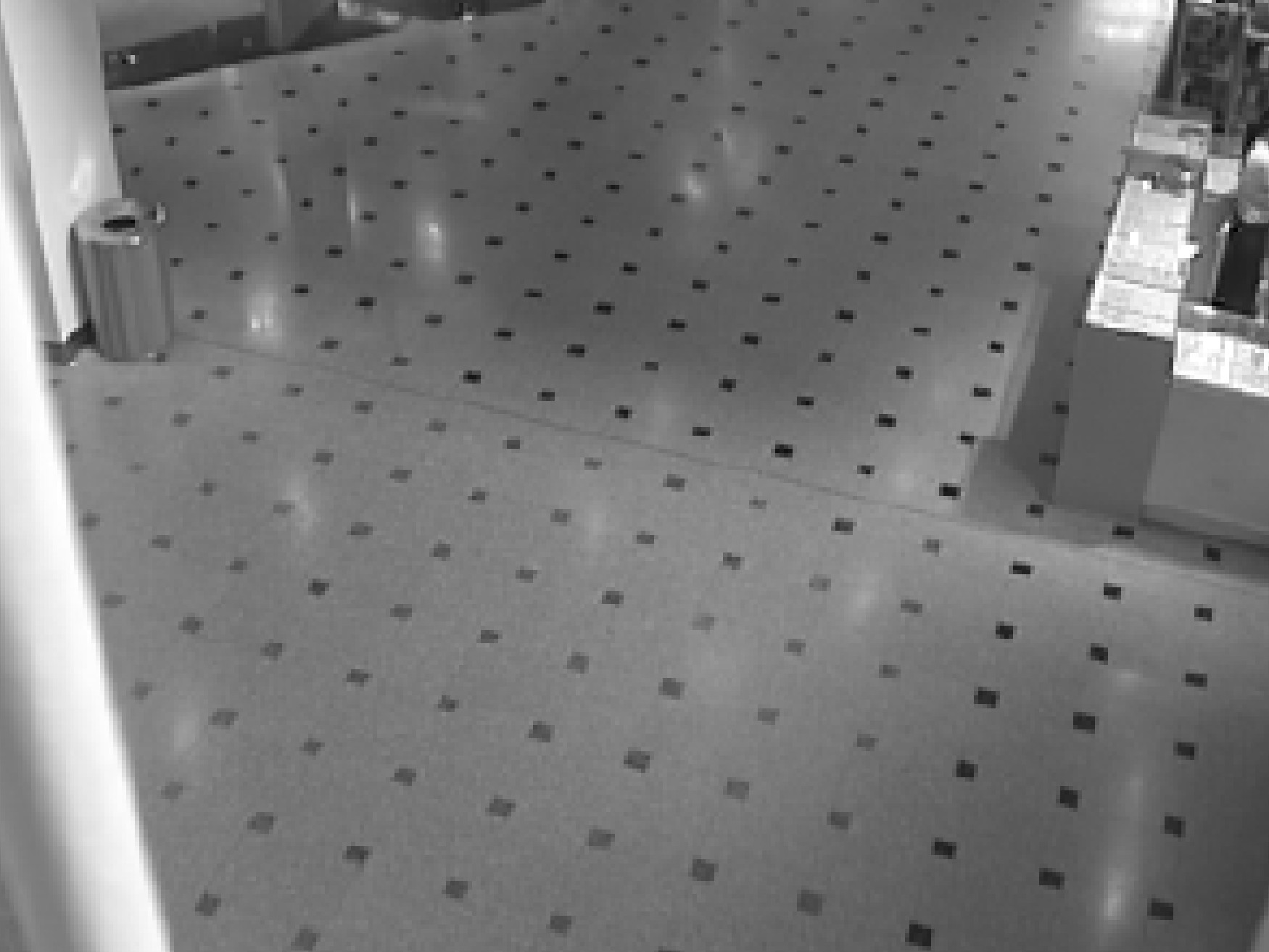}}\hfill 
\subfloat{\includegraphics[width=.245\linewidth]{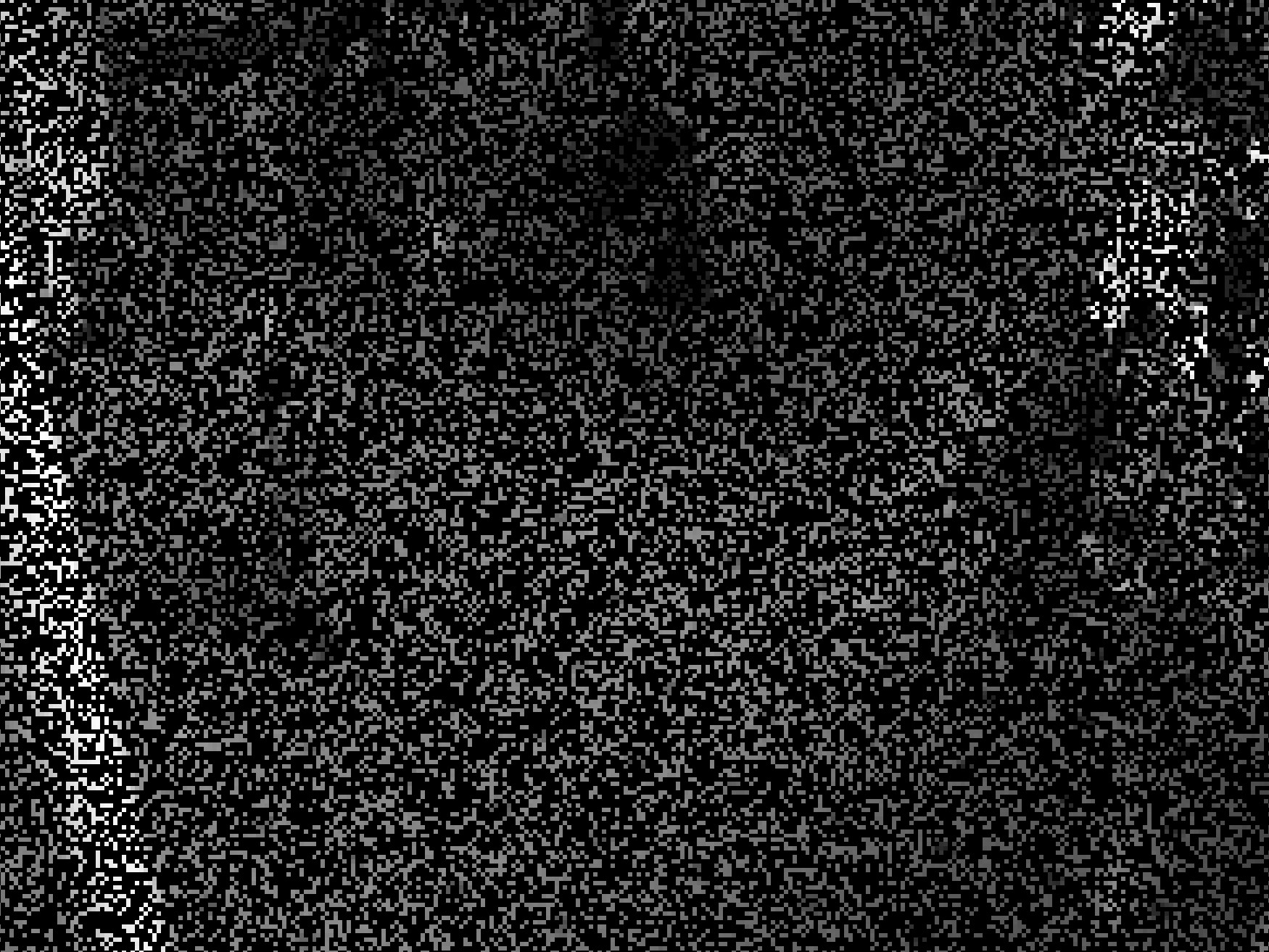}}\hfill
\subfloat{\includegraphics[width=.245\linewidth]{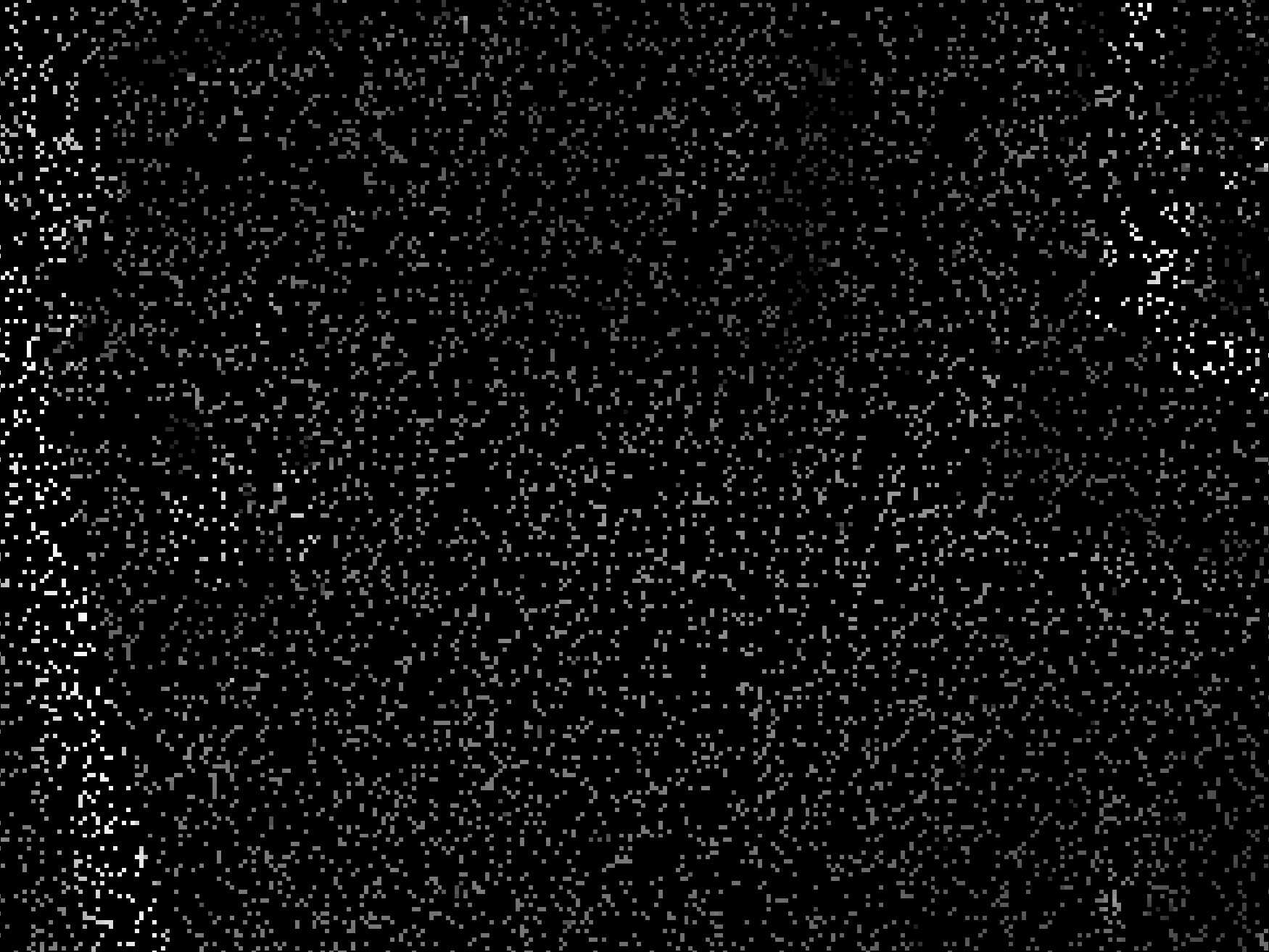}}\hfill
\subfloat{\includegraphics[width=.245\linewidth]{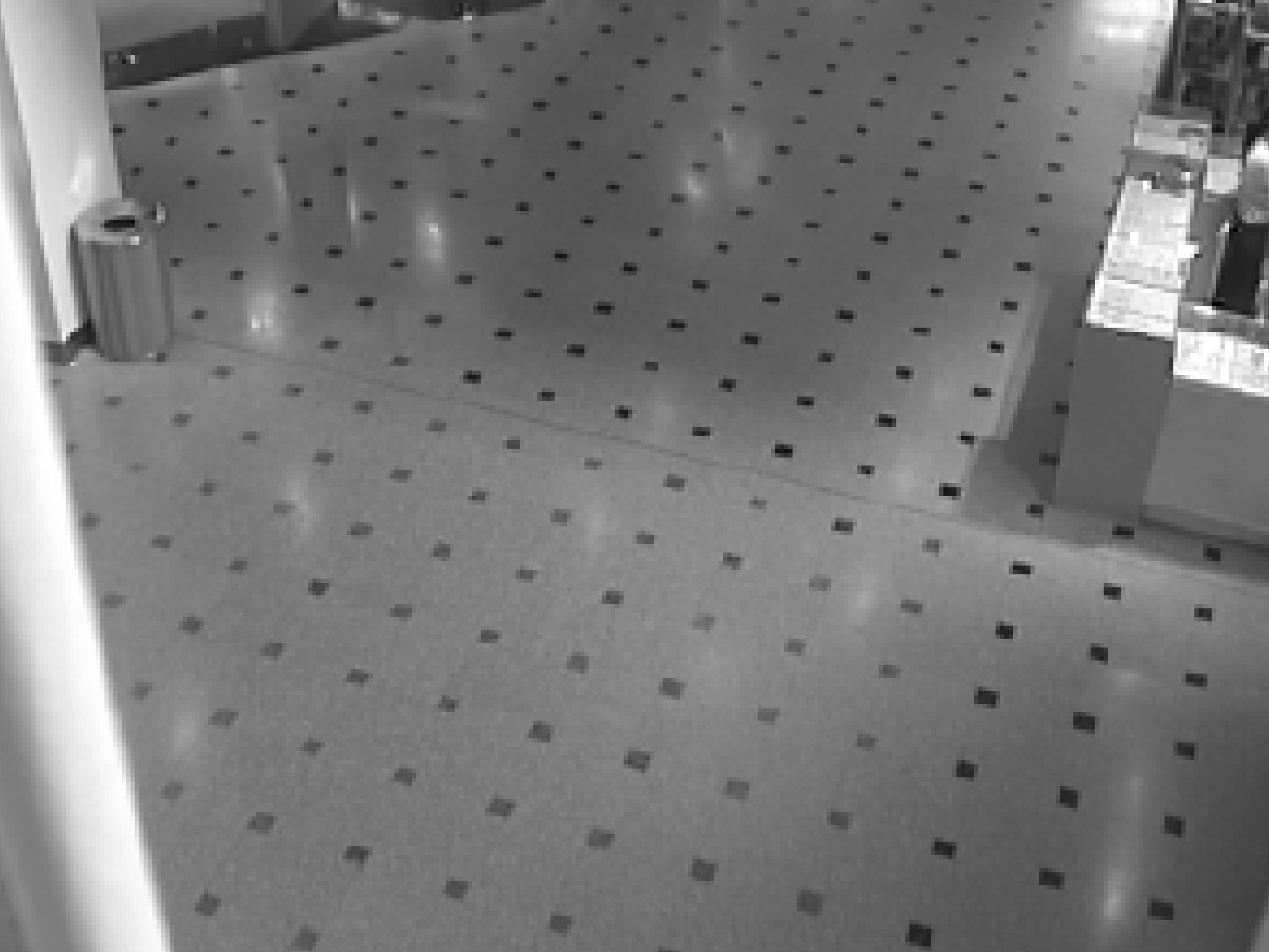}} 

\caption{
Video background subtraction results: The first row shows \textit{shoppingmall}, the second \textit{restaurant}. Column one displays the ground truth background via IRCUR from full datasets. Columns two and three show two observed frames from our CCS model. The last column presents the background from RCURC using partial videos. }\label{FIG:video background subtraction}
\vspace{-0.05in}
\end{figure}

To evaluate the performance of our method, we apply IRCUR  \cite{cai2020rapid} to the entire video dataset. We use the background derived from IRCUR as the benchmark ground truth. Subsequently, we randomly select \(40\%\) of the rows and \(40\%\) of the columns to create subrow and subcolumn matrices. At the next stage, we generate \(30\%\) observations on these submatrices. The reconstructed backgrounds obtained using RCURC are illustrated in \Cref{FIG:video background subtraction}. 
It is evident that the background recovered by RCURC closely resembles that obtained via IRCUR. Furthermore, we utilize the Peak Signal-to-Noise Ratio (PSNR) as a quantitative measure to evaluate the reconstruction quality of RCURC in comparison to the results achieved with IRCUR. These findings are detailed in  \Cref{tab:video}.

\begin{table}[t]
\caption{Video size, runtime and PSNR. Herein $\mathcal{S}$ represents \textit{shoppingmall} and $\mathcal{R}$ represents \textit{restaurant}.}\label{tab:video}
 \centering
\tabcolsep=0.15cm
 \begin{tabular}{ c|c|c|c|c} 
\toprule
 ~             &\textsc{Frame Size} & \textsc{Frame Number} &    \textsc{Runtime} & PSNR \cr
\midrule

$\mathcal{S}$  &$256\times 320$              & $1000$                     &   $33.12$s &  $41.87$        \cr

$\mathcal{R}$   & $120\times 160$                        &   $3055$   &  $31.16$s    &  $39.84$ \cr
\bottomrule
\end{tabular}
\end{table}
 
\section{Conclusion Remarks} \label{sec:conclusion}
This paper introduces a novel mathematical model for robust matrix completion problems with cross-concentrated samples. A highly efficient non-convex algorithm, dubbed RCURC, has been developed for the proposed model. The key techniques are projected gradient descent and CUR approximation. The numerical experiments, with both synthetic and real datasets, show high potential. In particular, we consistently observe linear convergence on RCURC. 

As for future work, we will study the statistical properties of the proposed robust CCS completion model, such as theoretical sample complexities and outlier tolerance. The recovery guarantee with a linear convergence rate will also be established for RCURC. We will also explore other real-world applications that suit the proposed model.

\bibliographystyle{IEEEtran}
\bibliography{IEEEabrv,ref}

\begin{thebibliography}{10}
\providecommand{\url}[1]{#1}
\csname url@samestyle\endcsname
\providecommand{\newblock}{\relax}
\providecommand{\bibinfo}[2]{#2}
\providecommand{\BIBentrySTDinterwordspacing}{\spaceskip=0pt\relax}
\providecommand{\BIBentryALTinterwordstretchfactor}{4}
\providecommand{\BIBentryALTinterwordspacing}{\spaceskip=\fontdimen2\font plus
\BIBentryALTinterwordstretchfactor\fontdimen3\font minus
  \fontdimen4\font\relax}
\providecommand{\BIBforeignlanguage}[2]{{%
\expandafter\ifx\csname l@#1\endcsname\relax
\typeout{** WARNING: IEEEtran.bst: No hyphenation pattern has been}%
\typeout{** loaded for the language `#1'. Using the pattern for}%
\typeout{** the default language instead.}%
\else
\language=\csname l@#1\endcsname
\fi
#2}}
\providecommand{\BIBdecl}{\relax}
\BIBdecl

\bibitem{candes2009exact}
E.~J. Cand{\`e}s and B.~Recht, ``Exact matrix completion via convex
  optimization,'' \emph{Found. Comput. Math.}, vol.~9, no.~6, pp. 717--772,
  2009.

\bibitem{recht2011simpler}
B.~Recht, ``A simpler approach to matrix completion.'' \emph{J. Mach. Learn.
  Res.}, vol.~12, no.~12, 2011.

\bibitem{Netflix}
J.~Bennett, C.~Elkan, B.~Liu, P.~Smyth, and D.~Tikk, ``{KDD} cup and workshop
  2007,'' \emph{SIGKDD Explor. Newsl.}, vol.~9, no.~2, p. 51–52, 2007.

\bibitem{goldberg1992using}
D.~Goldberg, D.~Nichols, B.~M. Oki, and D.~Terry, ``Using collaborative
  filtering to weave an information tapestry,'' \emph{Commun. ACM}, vol.~35,
  no.~12, pp. 61--70, 1992.

\bibitem{chen2004recovering}
P.~Chen and D.~Suter, ``Recovering the missing components in a large noisy
  low-rank matrix: Application to sfm,'' \emph{IEEE Trans. Pattern Anal. Mach.
  Intell.}, vol.~26, no.~8, pp. 1051--1063, 2004.

\bibitem{hu2012fast}
Y.~Hu, D.~Zhang, J.~Ye, X.~Li, and X.~He, ``Fast and accurate matrix completion
  via truncated nuclear norm regularization,'' \emph{IEEE Trans. Pattern Anal.
  Mach. Intell.}, vol.~35, no.~9, pp. 2117--2130, 2012.

\bibitem{cai2019fast}
J.-F. Cai, T.~Wang, and K.~Wei, ``Fast and provable algorithms for spectrally
  sparse signal reconstruction via low-rank {H}ankel matrix completion,''
  \emph{Appl. Comput. Harmon. Anal.}, vol.~46, no.~1, pp. 94--121, 2019.

\bibitem{cai2023hsgd}
H.~Cai, J.-F. Cai, and J.~You, ``Structured gradient descent for fast robust
  low-rank {H}ankel matrix completion,'' \emph{SIAM J. Sci. Comput.}, vol.~45,
  no.~3, pp. A1172--A1198, 2023.

\bibitem{cai2023ccs}
H.~Cai, L.~Huang, P.~Li, and D.~Needell, ``Matrix completion with
  cross-concentrated sampling: Bridging uniform sampling and {CUR} sampling,''
  \emph{IEEE Trans. Pattern Anal. Mach. Intell.}, vol.~45, no.~8, pp.
  10\,100--10\,113, 2023.

\bibitem{candes2011robust}
E.~J. Cand{\`e}s, X.~Li, Y.~Ma, and J.~Wright, ``Robust principal component
  analysis?'' \emph{Journal of the ACM}, vol.~58, no.~3, pp. 1--37, 2011.

\bibitem{netrapalli2014non}
P.~Netrapalli, U.~Niranjan, S.~Sanghavi, A.~Anandkumar, and P.~Jain,
  ``Non-convex robust {PCA},'' in \emph{Adv. Neural Inf. Process Syst.}, 2014,
  pp. 1107--1115.

\bibitem{yi2016fast}
X.~Yi, D.~Park, Y.~Chen, and C.~Caramanis, ``Fast algorithms for robust {PCA}
  via gradient descent,'' in \emph{Adv. Neural Inf. Process Syst.}, 2016, pp.
  4152--4160.

\bibitem{zhang2018robust}
T.~Zhang and Y.~Yang, ``Robust pca by manifold optimization,'' \emph{J. Mach.
  Learn. Res.}, vol.~19, no.~1, pp. 3101--3139, 2018.

\bibitem{cai2019accelerated}
H.~Cai, J.-F. Cai, and K.~Wei, ``Accelerated alternating projections for robust
  principal component analysis,'' \emph{J. Mach. Learn. Res.}, vol.~20, no.~1,
  pp. 685--717, 2019.

\bibitem{tong2021accelerating}
T.~Tong, C.~Ma, and Y.~Chi, ``Accelerating ill-conditioned low-rank matrix
  estimation via scaled gradient descent,'' \emph{J. Mach. Learn. Res.},
  vol.~22, no. 150, pp. 1--63, 2021.

\bibitem{cai2021lrpca}
H.~Cai, J.~Liu, and W.~Yin, ``Learned robust {PCA}: A scalable deep unfolding
  approach for high-dimensional outlier detection,'' in \emph{Adv. Neural Inf.
  Process Syst.}, vol.~34, 2021, pp. 16\,977--16\,989.

\bibitem{chen2014robust}
Y.~Chen and Y.~Chi, ``Robust spectral compressed sensing via structured matrix
  completion,'' \emph{IEEE Trans. Inf. Theory}, vol.~60, no.~10, pp.
  6576--6601, 2014.

\bibitem{zhang2019correction}
S.~Zhang and M.~Wang, ``Correction of corrupted columns through fast robust
  hankel matrix completion,'' \emph{IEEE Trans. Signal Process.}, vol.~67,
  no.~10, pp. 2580--2594, 2019.

\bibitem{cai2021accelerated}
H.~Cai, J.-F. Cai, T.~Wang, and G.~Yin, ``Accelerated structured alternating
  projections for robust spectrally sparse signal recovery,'' \emph{IEEE Trans.
  Signal Process.}, vol.~69, pp. 809--821, 2021.

\bibitem{cai2021ircur}
H.~Cai, K.~Hamm, L.~Huang, J.~Li, and T.~Wang, ``Rapid robust principal
  component analysis: {CUR} accelerated inexact low rank estimation,''
  \emph{IEEE Signal Process. Lett.}, vol.~28, pp. 116--120, 2021.

\bibitem{cai2021rcur}
H.~Cai, K.~Hamm, L.~Huang, and D.~Needell, ``Robust {CUR} decomposition: Theory
  and imaging applications,'' \emph{SIAM J. Imaging Sci.}, vol.~14, no.~4, pp.
  1472--1503, 2021.

\bibitem{hamm2022RieCUR}
K.~Hamm, M.~Meskini, and H.~Cai, ``Riemannian {CUR} decompositions for robust
  principal component analysis,'' in \emph{Topological, Algebraic and Geometric
  Learning Workshops}, 2022, pp. 152--160.

\bibitem{HH2020}
K.~Hamm and L.~Huang, ``Perspectives on {CUR} decompositions,'' \emph{Appl.
  Comput. Harmon. Anal.}, vol.~48, no.~3, pp. 1088--1099, 2020.

\bibitem{chiu2013sublinear}
J.~Chiu and L.~Demanet, ``Sublinear randomized algorithms for skeleton
  decompositions,'' \emph{SIAM J. Matrix Anal. Appl.}, vol.~34, no.~3, pp.
  1361--1383, 2013.

\bibitem{chaturantabut2010nonlinear}
S.~Chaturantabut and D.~C. Sorensen, ``Nonlinear model reduction via discrete
  empirical interpolation,'' \emph{SIAM J. Sci. Comput.}, vol.~32, no.~5, pp.
  2737--2764, 2010.

\bibitem{SorensenDEIMCUR}
D.~C. Sorensen and M.~Embree, ``A {DEIM} induced {CUR} factorization,''
  \emph{SIAM J. Sci. Comput.}, vol.~38, no.~3, pp. A1454--A1482, 2016.

\bibitem{DMM08}
P.~Drineas, M.~W. Mahoney, and S.~Muthukrishnan, ``Relative-error {CUR} matrix
  decompositions,'' \emph{SIAM J. Matrix Anal. Appl.}, vol.~30, no.~2, pp.
  844--881, 2008.

\bibitem{HH2020S}
K.~Hamm and L.~Huang, ``Stability of sampling for {CUR} decompositions,''
  \emph{Found. Data Sci.}, vol.~2, no.~2, pp. 83--99, 2020.

\bibitem{cai2021rtcur}
H.~Cai, Z.~Chao, L.~Huang, and D.~Needell, ``Fast robust tensor principal
  component analysis via fiber {CUR} decomposition,'' in \emph{Proceedings of
  the IEEE/CVF International Conference on Computer Vision Workshops}, 2021,
  pp. 189--197.

\bibitem{cai2024rtcur}
------, ``Robust tensor {CUR} decompositions: Rapid low-tucker-rank tensor
  recovery with sparse corruption,'' \emph{SIAM J. Imaging Sci.}, 2024.

\bibitem{cai2020rapid}
H.~Cai, K.~Hamm, L.~Huang, J.~Li, and T.~Wang, ``Rapid robust principal
  component analysis: {CUR} accelerated inexact low rank estimation,''
  \emph{IEEE Signal Process. Lett.}, vol.~28, pp. 116--120, 2020.

\end{thebibliography}
\end{document}